\documentclass[10pt,twocolumn,letterpaper]{article}

\usepackage{wacv}              

\usepackage{graphicx}
\usepackage{amsmath}
\usepackage{amssymb}
\usepackage{booktabs}
\usepackage{array,makecell}
\usepackage{pifont}
\newcommand{\cmark}{\ding{51}}%
\newcommand{\xmark}{\ding{55}}%
\usepackage{color,soul}

\usepackage[pagebackref,breaklinks,colorlinks]{hyperref}

\usepackage[capitalize]{cleveref}
\crefname{section}{Sec.}{Secs.}
\Crefname{section}{Section}{Sections}
\Crefname{table}{Table}{Tables}
\crefname{table}{Tab.}{Tabs.}

\def\wacvPaperID{947} 
\def\confName{WACV}
\def\confYear{2024}

\begin{document}

\title{Separable Self and Mixed Attention Transformers for Efficient Object Tracking}

\author{Goutam Yelluru Gopal \hspace{80pt} Maria A. Amer \\
	{\tt\small g\_yellur@encs.concordia.ca} \hspace{45pt} {\tt\small amer@ece.concordia.ca} \\
Department of Electrical and Computer Engineering, Concordia University, Montr{\'e}al, Qu{\'e}bec, Canada
}
\maketitle

\begin{abstract}
The deployment of transformers for visual object tracking has shown state-of-the-art results on several benchmarks. However, the transformer-based models are under-utilized for Siamese lightweight tracking due to the computational complexity of their attention blocks. This paper proposes an efficient self and mixed attention transformer-based architecture for lightweight tracking. The proposed backbone utilizes the separable mixed attention transformers to fuse the template and search regions during feature extraction to generate superior feature encoding. Our prediction head performs global contextual modeling of the encoded features by leveraging efficient self-attention blocks for robust target state estimation. With these contributions, the proposed lightweight tracker deploys a transformer-based backbone and head module concurrently for the first time. Our ablation study testifies to the effectiveness of the proposed combination of backbone and head modules. Simulations show that our Separable Self and Mixed Attention-based Tracker, \textit{SMAT}, surpasses the performance of related lightweight trackers on GOT10k, TrackingNet, LaSOT, NfS30, UAV123, and AVisT datasets, while running at 37 \textit{fps} on CPU, 158 fps on GPU, and having 3.8M parameters. For example, it significantly surpasses the closely related trackers E.T.Track and MixFormerV2-S on GOT10k-test by a margin of 7.9\% and 5.8\%, respectively, in the $AO$ metric. The tracker code and model is available at \url{https://github.com/goutamyg/SMAT}
\end{abstract}

\section{Introduction} 
\label{sec:intro}
The Siamese Network-based (SN) architecture is prevalent in visual object tracking due to its simplicity and high speed \cite{zhu2018distractor, yan2021learning}. The SN architecture consists of a backbone to generate robust feature representation of the target template and search regions, a localization head module for target state estimation, and an optional feature fusor module for relation modeling \cite{ye2022joint}. In recent years, the transformer-based \cite{vaswani2017attention, dosovitskiy2021an, wu2021cvt} tracking methods have unified feature extraction and relation modeling by deploying self and mixed attention blocks in their backbone \cite{yan2021learning, cui2022mixformer} to simplify the SN architecture further. Enabled by the computational power of GPUs, these transformer-based SN trackers achieve high frames-per-second (\textit{fps}) during inference. However, the high computational complexity of these transformer-based tracking algorithms severely impacts the \textit{fps} on constrained hardware, \textit{e.g.}, CPUs, limiting the utility of these algorithms towards applications with hardware constraints. 

The lightweight SN-based trackers \cite{borsuk2022fear, yan2021lighttrack, blatter2023efficient}, which are specifically designed for resource-constrained environments, adopt efficient building blocks to maintain real-time speed, i.e., $\ge 30$ \textit{fps}. Therefore, these trackers cannot fully leverage the modeling power of transformers in their architecture because of the computational complexity of the standard transformers, especially the expensive matrix multiplication while computing attention. Mehta \textit{et al.} \cite{mehta2023separable} addressed this issue by replacing the costly matrix-to-matrix multiplication with separable elementwise operations to present an efficient Mobile Vision Transformer (ViT) block for vision-related tasks. Leveraging these advances, we propose a separable self and mixed attention transformer-based lightweight architecture for real-time tracking.

\begin{figure}[t]
	\centering
	\includegraphics[width=\linewidth]{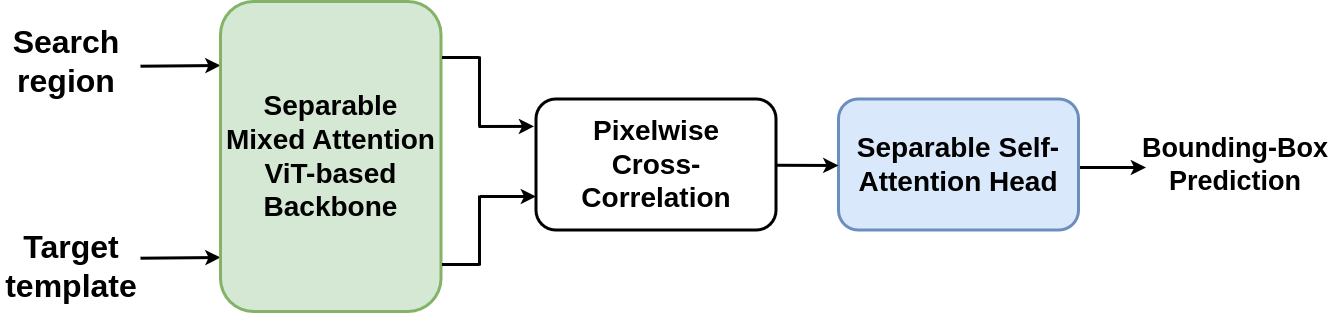}
	
	\caption{Proposed \textit{SMAT} architecture. The separable mixed attention Vision Transformer-based backbone jointly performs feature extraction and fusion of template and search regions. The separable transformer-based head models long-range dependencies within the fused features to predict accurate bounding boxes.}
	\label{fig:smat_architecture}
\end{figure}
The architecture of the proposed tracker is shown in Figure \ref{fig:smat_architecture}. We employ a cascaded arrangement of convolutional neural network (CNN) and ViT blocks in the proposed tracker backbone. Such a hybrid design \cite{mehta2022mobilevit} combines the merits of convolutions (i.e., learning the spatially-local representations) and transformers (i.e., modeling the long-range dependencies) with fewer parameters compared to the fully transformer-based backbone architecture, such as \cite{cui2022mixformer, ye2022joint}. Apart from generating a robust feature representation, the proposed backbone facilitates the exchange of information between the target template and search region by computing mixed attention \cite{cui2022mixformer} in the ViT block without bloating the backbone latency. Our prediction head efficiently performs global contextual modeling of encoded features using separable self-attention units. Such transformer-based global modeling of encoded features improves the localization accuracy compared to the fully convolution-based methods, as shown in \cite{blatter2023efficient}. With these contributions, we propose a lightweight self and mixed attention transformers-based tracker, \textit{SMAT}, running beyond real-time speed on a CPU.

\section{Related Work}
\label{sec:related_work}
The introduction of transformer-based modeling has significantly improved the performance of SN-based trackers in recent years \cite{9913708}. These SN trackers \cite{wang2021transformer, yan2021learning, zhao2021trtr, chen2021transformer} exploited the global contextual modeling capabilities of the transformer layers for relation modeling \cite{ye2022joint}, i.e., to fuse the features extracted from the target template and the search region. The deployment of computationally expensive transformer-based backbones for SN tracking \cite{ye2022joint, cui2022mixformer, lin2022swintrack, Wei_2023_CVPR, chen2023seqtrack, gao2023generalized} has improved the tracker performance further and achieved state-of-the-art results on various challenging benchmarks \cite{Huang2021, muller2018trackingnet, fan2021lasot}. 

In recent years, there have been several SN-based lightweight algorithms proposed for efficient object tracking. LightTrack \cite{yan2021lighttrack} used neural architectural search \cite{chen2019detnas} to design an efficient backbone and head modules suitable for resource-constrained environments. FEAR \cite{borsuk2022fear} presented a compact and energy-efficient SN-based tracking method running at real-time speed on a smartphone. It uses the dual-template representation with a dynamic update scheme to model the target appearance variations. Stark-Lightning \cite{yan2021learning} proposed an efficient tracking method with a lightweight transformer-based feature fusor module. HiFT \cite{cao2021hift} used hierarchical feature transformers to achieve real-time speed on an embedded processor for aerial tracking. SiamHFFT \cite{dai2022siamese} introduced a hierarchical transformer-based feature fusion module for efficient tracking on CPUs. HCAT \cite{chen2022efficient} deployed a feature sparsification module and a hierarchical cross-attention transformer-based architecture to achieve real-time on edge devices. It should be noted that the transformer-based tracker \cite{yan2021learning, cao2021hift, dai2022siamese, chen2022efficient} employ CNN-based backbones for feature extraction and utilize the transformer layers only for relation modeling, i.e., to fuse the feature representations generated by their backbones.

Fewer lightweight SN trackers use transformer modules in their backbone or head architecture. MixFormerV2 \cite{cui2023mixformerv2} presented a fully transformer-based \cite{dosovitskiy2021an} backbone for efficient tracking, based on knowledge-distillation \cite{hinton2015distilling} and progressive model-depth pruning. E.T.Track \cite{blatter2023efficient} proposed an efficient Exemplar Transformer-based prediction head for visual tracking. It utilized a single instance-level attention layer in the transformer block to achieve real-time on a CPU. Compared to the other related trackers, \cite{cui2023mixformerv2} and \cite{blatter2023efficient} are the closest to our work.

Unlike the two-stream encoding approach by \cite{borsuk2022fear, cao2021hift, dai2022siamese, chen2022efficient, blatter2023efficient, yan2021learning, yan2021lighttrack} (i.e., the template and search region features are extracted independently), the proposed backbone facilitates the exchange of information between template and search regions during feature extraction. Different from \cite{borsuk2022fear, cao2021hift, dai2022siamese, chen2022efficient, cui2023mixformerv2, yan2021learning}, we use a transformer-based head for target localization. In contrast to the fully transformer-based backbone by \cite{cui2023mixformerv2}, we use a cascade of CNN \cite{sandler2018mobilenetv2} and ViT \cite{mehta2023separable} blocks in our backbone. Also, the iterative nature of knowledge-distillation and model pruning by \cite{cui2023mixformerv2} requires multiple rounds of model training, whereas our tracker model needs to be trained only once. Compared to the exemplar transformer-based head module by E.T.Track \cite{blatter2023efficient}, our separable self-attention prediction head is $3 \times$ compact in terms of parameters (5.7 Million for E.T.Track versus 1.8 Million for our \textit{SMAT}). As a post-processing step, the related \cite{yan2021lighttrack} and \cite{blatter2023efficient} refine the predicted bounding boxes by penalizing large changes in target size and aspect ratio of the predicted boxes between consecutive frames. We do not employ any bounding-box refinement techniques during post-processing.

To summarize our contributions, we are the first to propose:
\begin{itemize}
	\item A separable mixed attention ViT-based backbone for joint feature extraction and information fusion between the template and search regions. Our approach combines the merits of CNNs and efficient ViTs to learn superior feature encoding for accurate tracking without bloating the backbone latency.
	\item A separable self-attention transformer-based prediction head to model the global dependencies within the fused feature encoding for accurate bounding-box prediction.
\end{itemize} 
Compared to related work, for the first time, our method concurrently deploys an efficient transformer-based backbone and prediction head for lightweight tracking.

\section{The Proposed Method}
\label{sec:method}
This section discusses the architecture and training details of the proposed \textit{SMAT} tracker. Section \ref{subsec:backbone} presents our tracker backbone and Section \ref{subsec:head} describes the architecture of the proposed head module. Section \ref{subsec:loss} has details of the loss function used during training. 

\begin{figure*}[t]
	\centering
	\includegraphics[width=\linewidth]{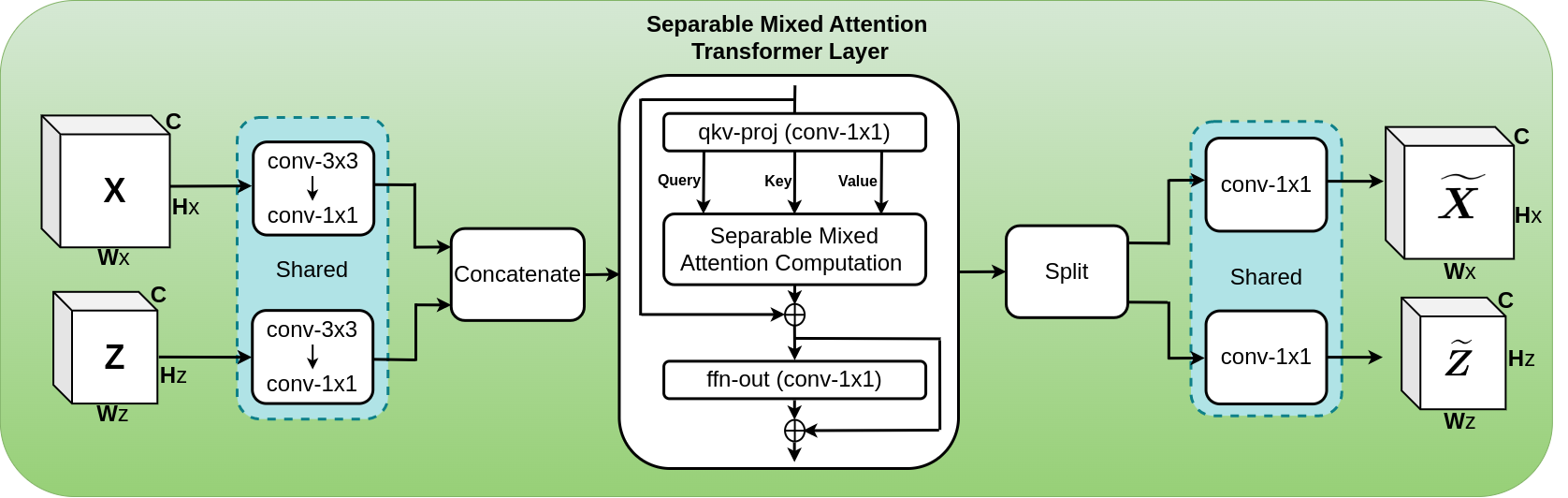}
	
	\caption{Proposed Separable Mixed Attention ViT block. The \textit{qkv-proj} denotes the set of three $1 \times 1$ convolutional filters to generate the \textit{Query}, \textit{Key}, and \textit{Value} for attention computation. The mixed attention output is passed through a $1 \times 1$ convolutional \textit{ffn-out} block to generate the output of the transformer layer.}
	\label{fig:smat_backbone}
\end{figure*}
\subsection{Proposed Mixed Attention ViT-based Backbone}
\label{subsec:backbone}
The backbone of the proposed tracker receives two images as its input; one is the target template $Z_{in}$, and the other is the search region for target localization, $X_{in}$. First, we apply the CNN-based Inverted Residual (IR) \cite{sandler2018mobilenetv2} blocks on $Z_{in}$ and $X_{in}$. These IR blocks generate spatially local feature representations of $Z_{in}$ and $X_{in}$ while being efficient compared to the regular convolutional blocks \cite{sandler2018mobilenetv2}. In addition, these IR blocks reduce the spatial dimensionality of input images by the pooling operation to generate low-dimensional feature representations for our separable mixed attention ViT block. 

The architecture of the proposed mixed attention ViT block is shown in Figure \ref{fig:smat_backbone}. Let $Z \in R^{W_z \times H_z \times C}$ and $X \in R^{W_x \times H_x \times C}$ denote the template and search region feature representations, respectively. Inside the proposed ViT block, we initially pass $Z$ and $X$ through a series of $3 \times 3$ and $1 \times 1$ CNN layers with shared weights to project the number of channels in $Z$ and $X$ from $C$ to $d$. We tokenize \cite{mehta2022mobilevit} the output of CNN blocks and concatenate them to generate a total of $k$ tokens to learn mixed attention \cite{cui2022mixformer} between the template and search regions. Inside the transformer layer, we first apply a set of three $1 \times 1$ convolutional filters (denoted as \textit{qkv-proj} in Figure \ref{fig:smat_backbone}) to generate the query $\mathcal{Q} \in R^{k \times 1}$, the key $\mathcal{K} \in R^{k \times d}$, and the value $\mathcal{V} \in R^{k \times d}$. Then, we apply the softmax operation on the query vector $\mathcal{Q}$ and broadcast along its column (i.e., the element in $i^{th}$ row is repeated $d$ times along the column dimension) to generate $\mathcal{\tilde{Q}} \in R^{k \times d}$. Using $\mathcal{\tilde{Q}}$ and $\mathcal{K}$, the context vector $\mathcal{A} \in R^{1 \times d}$ is computed as
\begin{equation}
	\mathcal{A} = \sum_{k} \mathcal{\tilde{Q}} \odot \mathcal{K},
	\label{eq:compute_attn}
\end{equation}
where $\odot$ denotes the element-wise multiplication and $\sum_{k}$ indicates summation across the rows. The context vector $\mathcal{A}$ is broadcasted along its rows to create $\mathcal{\tilde{A}} \in R^{k \times d}$, which is used to compute the mixed attention $\mathcal{M} \in R^{k \times d}$ as  
\begin{equation}
	\mathcal{M} = \mathcal{\tilde{A}} \odot \text{ReLU}(\mathcal{V}).
	\label{eq:compute_attn_2}
\end{equation}
The elementwise multiplication operations in Eq. \ref{eq:compute_attn} and Eq. \ref{eq:compute_attn_2} reduce the latency of the separable transformer layer, shown in Figure \ref{fig:smat_backbone}, when compared to the dense matrix-to-matrix multiplication-based attention computation in standard transformers \cite{vaswani2017attention}. Also, computing the mixed attention on the concatenated features concurrently models the global interactions \textit{within} (i.e., self) and \textit{between} (i.e., cross) the target template and the search area. Therefore, mixed attention requires fewer transformer block evaluations than separately computing the self and cross-attention. Similar to \cite{vaswani2017attention}, we employ a residual connection \cite{he2016deep} around the attention computation block. We pass the output of the residual connection through a $1 \times 1$ convolutional feedforward network (denoted as \textit{ffn-out} in Figure \ref{fig:smat_backbone}) to generate the output of the separable transformer layer. We split the resulting feature map to separate the template and search region features with $d$ channels. Finally, we re-project the number of channels from $d$ to $C$ by applying a shared $1 \times 1$ convolutional filter on the separated feature maps to generate $\tilde{X}$ and $\tilde{Z}$ as the output. 

The computation of mixed attention in the ViT block facilitates implicit relation modeling during feature extraction, thereby generating superior features compared to the two-stream encoding approach. Such feature fusion also avoids needing a parameter-heavy module for the subsequent relation modeling between the template and search region features. For a parameter-efficient relation modeling (or feature fusion) between the features $\tilde{X}$ and $\tilde{Z}$ generated by the proposed backbone, we use the \textit{parameter-free} pixel-wise cross-correlation \cite{yan2021alpha} operation (\textit{cf.} Figure \ref{fig:smat_architecture}). The fused feature encoding $F$ is computed as,
\begin{equation}
	F = PWCorr(\tilde{X}, \tilde{Z}).
	\label{eq:pw_xcorr}
\end{equation}
where $PWCorr$ denotes the pixel-wise cross-correlation operation. We apply a $1 \times 1$ convolution filter on the fused encoding to transform the number of channels to $C_h$. To perform target localization, we pass the resulting encoding $F$ to the proposed prediction head in Section \ref{subsec:head}.    

\subsection{Proposed Self-Attention Prediction Head}\label{subsec:predictor_head}
\begin{figure}[t]
	\centering
	\includegraphics[width=\linewidth]{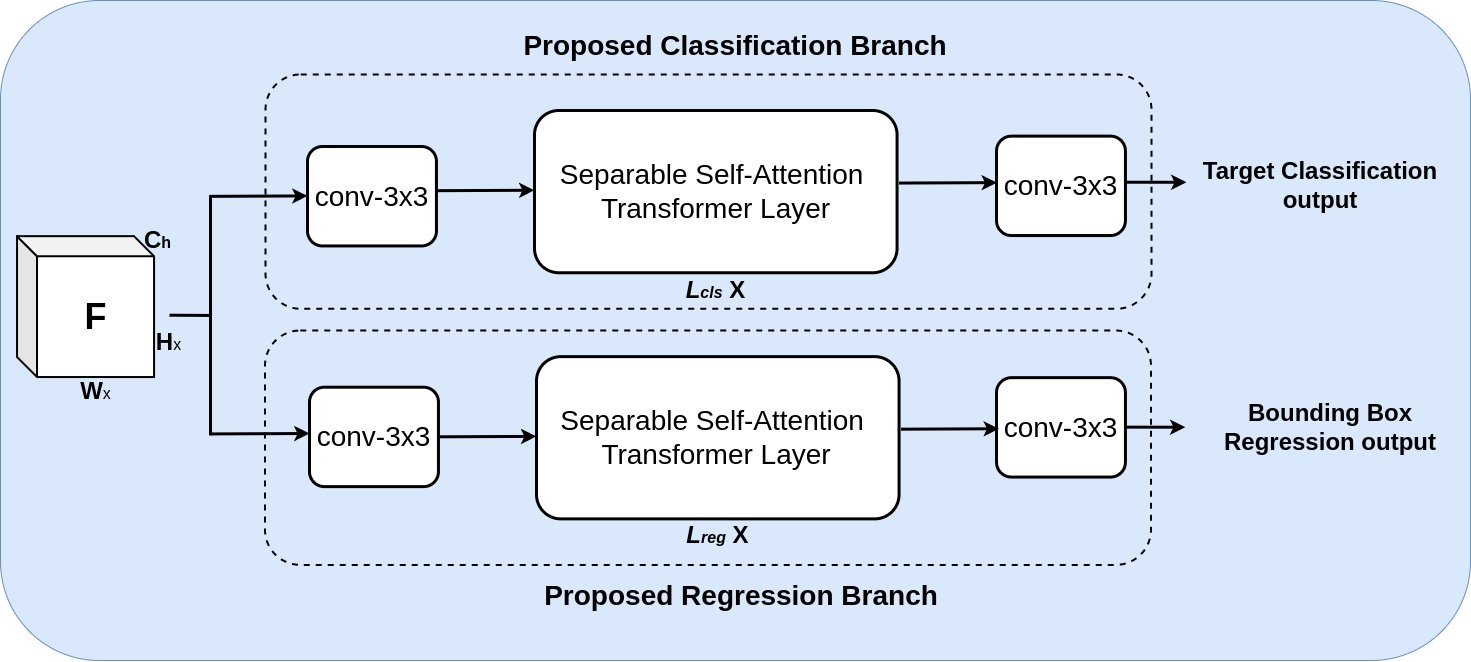}
	\caption{Proposed Separable Self-Attention Transformer-based Predictor Head. It utilizes two branches for target classification and bounding-box regression.}
	\label{fig:smat_head}
\end{figure}

The pipeline of the proposed predictor head is shown in Figure \ref{fig:smat_head}. Our prediction head has two branches: target classification and bounding-box regression. Unlike a fully CNN-based approach, we implement these branches via a convolutional and transformer layers cascade. The CNN layers are suitable for modeling the local relationships within the fused feature representation but are limited in effectively capturing the non-local associations. On the other hand, the transformer layers explicitly model the long-range global interactions within the features by processing the tokenized feature encoding. Such a global modeling scheme is beneficial for localization under scenarios of drastic target shape variations and heavy occlusion \cite{yan2021learning}. The cascade of convolutional and transformer-based layers combines the best of their modeling strengths to produce high-quality tracking results on challenging videos.
\label{subsec:head}

First, we apply a $3 \times 3$ convolutional filter for the proposed predictor head on the fused encoding $F$ from Eq. \ref{eq:pw_xcorr} to extract spatially local feature representations. We then tokenize the filter output and pass it through a stack of $L_{cls}$ and $L_{reg}$ separable self-attention transformer layers in classification and regression branches, respectively. For these transformer blocks, the pipeline of computing the attention is similar to the transformer layer in the ViT block from Section \ref{subsec:backbone}; except, here, we calculate self-attention to model long-range dependencies \textit{within} the fused feature encoding $F$ for robust target state estimation. The output of the self-attention transformer layer is passed through a $3 \times 3$ CNN layer to generate a score map $\mathcal{R}$ for the classification branch, and the local offset and the normalized bounding-box size for the regression branch, as in \cite{ye2022joint}. A combination of local and long-range contextual modeling by our predictor head improves tracker performance without significantly increasing the overall model latency.

\subsection{Loss function for Training}
\label{subsec:loss}
We use loss functions on the classification and regression output generated by the proposed head module while training our model. We use the weighted focal loss for the output of the classification branch; for the output of the regression branch, we use the $\ell_1$ loss and generalized Intersection-over-Union ($IoU$) loss, as in \cite{ye2022joint}. The overall loss function $\mathcal{L}_{total}$ is defined as
\begin{equation}
	\mathcal{L}_{total} = \mathcal{L}_{focal} + \lambda_{\ell_1} \cdot \mathcal{L}_{\ell_1} + \lambda_{IoU} \cdot \mathcal{L}_{IoU},
	\label{eq:total_loss}
\end{equation}
where $\mathcal{L}_{focal}$, $\mathcal{L}_{\ell_1}$, and $\mathcal{L}_{IoU}$ represent the focal loss, $\ell_1$ loss and the generalized $IoU$ loss functions, respectively. $\lambda_{\ell_1}$ and $\lambda_{IoU}$ are the hyperparameters determining the relative impact of the respective loss functions.

\section{Experimental and Ablation Results}
\subsection{Implementation Details}
We set the size of the template and search region images, i.e., $Z_{in}$ and $X_{in}$ from Section \ref{sec:method}, to $128 \times 128$ and $256 \times 256$, respectively, during training and inference. We deploy two CNN-based IR \cite{sandler2018mobilenetv2} blocks and two ViT-blocks in our backbone. The sequential ordering of the IR and ViT blocks is the same as MobileViTv2 \cite{mehta2023separable} pipeline. During feature extraction, our backbone performs four downsampling operations, each by a factor of 2; therefore, the spatial dimension of the features generated by our backbone is $8 \times 8$ and $16 \times 16$ for the template and search regions, respectively. For the proposed head module, we set the number of channels in the fused encoding, i.e., $C_h$ in Figure \ref{fig:smat_head}, to 128. We set the number of transformer layers $L_{cls}$ and $L_{reg}$ in the classification and regression branches to 2 and 4, respectively. The reason for defining $L_{reg}$ two times the value of $L_{cls}$ is because the regression head predicts twice the variables, i.e., local offset and bounding-box size, compared to the classification branch predicting the target center.

\subsection{Training and Inference Details}
We use the GOT10k \cite{Huang2021}, LaSOT \cite{fan2021lasot}, TrackingNet \cite{muller2018trackingnet}, and COCO \cite{lin2014microsoft} datasets to train our tracker. GOT10k, LaSOT, and TrackingNet have a non-overlapping train-test split ratio of 9335-to-180, 1120-to-280, and 30132-to-511 videos, respectively. Also, GOT10k provides 180 additional videos as the validation split. We apply data augmentation (horizontal flip and scale jittering) to generate training image pairs for the still images in the COCO train dataset. We use the combined training splits of these four datasets to train our model.

We train our model for 300 epochs, and each epoch uses $6 \times 10^4$ image pairs uniformly sampled from the training dataset. The initial learning rate (\textit{lr}) is set to 0.0004 and is reduced by a factor of 10 after 240 epochs. The \textit{lr} for the backbone parameters is set 0.1 times the \textit{lr} for the remaining trainable parameters of our model. We use AdamW \cite{loshchilov2018decoupled} as the network optimizer and set the weight decay to $10^{-4}$. The values of hyperparameters $\lambda_{\ell_1}$ and $\lambda_{IoU}$ from Eq. \ref{eq:total_loss} are set to 5 and 2, respectively. These hyperparameter values used for training are derived from \cite{ye2022joint} with no additional finetuning. We initialize our backbone weights using a pretrained MobileViTv2 model provided by the authors\cite{mehta2023separable}. We use PyTorch \cite{paszke2019pytorch} for developing the tracker code. Our model is trained on a single NVidia Telsa V100 GPU (32GB memory) with a batch size of 128. We monitor the possibility of overfitting by periodically evaluating the values of loss functions $\mathcal{L}_{focal}, \mathcal{L}_{\ell_1}$, and $\mathcal{L}_{IoU}$ from Eq. \ref{eq:total_loss} using the GOT10k validation videos.

During inference, we use the annotation from the first frame in the video as the target template and do not perform model update. To define the search region at frame $t$, we crop a region around tracker output at frame $t-1$, four times the target size. This image is resized to $256 \times 256$ and utilized as the search image at frame $t$. As a post-processing step, we apply a Hanning window on the classification score map $\mathcal{R}$ to penalize large target displacement predictions. 

\subsection{Comparison to the Related Work}
To assess the performance of the proposed \textit{SMAT}, we evaluate our tracker on the test-split of GOT10k \cite{Huang2021}, LaSOT \cite{fan2021lasot}, TrackingNet \cite{muller2018trackingnet}, NfS30 \cite{kiani2017need}, UAV123 \cite{mueller2016benchmark}, and AVisT \cite{noman2022avist} datasets. GOT10k-test has 180 test videos having non-overlapping object classes compared to their training videos, mainly to promote the generalization of tracking algorithms towards unseen object categories. LaSOT-test has 280 videos with 14 different attributes and balanced class categories. With an average length of 2500 frames per video, LaSOT-test is effective in accessing long-term tracking capabilities. TrackingNet-test contains 511 challenging videos curated from the large-scale YouTube-BB \cite{real2017youtube} dataset with 15 attribute annotations. NfS30 has 100 test videos, predominantly containing fast-moving objects with significant motion blur. UAV123 is a low-altitude UAV tracking benchmark and has 123 videos with 12 attribute annotations. AVisT dataset has 120 challenging videos with a wide range of atmospheric adverse scenarios such as rain, fog, fire, low-light, snow, tornado, and smoke impacting the target appearance in the test videos. The datasets NfS30, UAV123, and AVisT have no training split videos. 

We use the metrics recommended by the corresponding dataset authors to quantify the tracker performance during our evaluation. GOT10k uses the Average of Overlap ($AO$) based on the Intersection-of-Union (\textit{IoU}) value between the groundtruth and predicted bounding boxes, averaged over all the test videos. It also uses Success Rate ($SR$), computing the fraction of frames having an $IoU$ value greater than a threshold $\tau$, with values of $\tau$ as 0.5 and 0.75. TrackingNet uses Area-Under-the-Curve ($AUC$), Precision ($P$), and Normalized-Precision ($P_{norm}$) as the tracker evaluation metrics. $AUC$ is equivalent to $AO$ \cite{vcehovin2016visual}, and $P$ is computed based on the distance between the groundtruth and predicted bounding-box centers, measured in pixels. The metric $P_{norm}$ is similar to $P$; however, $P_{norm}$ uses normalized bounding boxes while measuring the distance between their centers. LaSOT uses the same evaluation metric as TrackingNet, whereas NfS30 and UAV123 use $AUC$ and $P$ for tracker evaluation. Along with the $AUC$, AVisT uses OP$50$ and OP$75$ as its evaluation metrics, which are equivalent to $SR$ at thresholds 0.5 and 0.75, respectively, from GOT10k. To ensure fair tracker evaluation and avoid finetuning of parameters on the test data, GOT10k and TrackingNet sequester the ground-truth annotations for their test videos. Therefore, we generate the metrics for these datasets by submitting the raw tracker results to the remote evaluation server. The groundtruth annotations are available for the LaSOT-test, NfS30, UAV123, and AVisT datasets. 

We compare the results of the proposed \textit{SMAT} against the related lightweight trackers: LightTrack \cite{yan2021lighttrack}, Stark-Lightning \cite{yan2021learning}, FEAR-XS \cite{borsuk2022fear}, HCAT \cite{chen2022efficient}, E.T.Track \cite{blatter2023efficient}, and MixFormerV2-S \cite{cui2023mixformerv2}, evaluated using the pretrained models provided by their authors. From Table \ref{table:qualitative_results}, we can see that the proposed \textit{SMAT} outperforms the related trackers on all six test datasets: GOT10k-test, TrackingNet-test, LaSOT-test, NfS30, UAV123, and AVisT. 
\begin{table*}[t]
	\centering
	\resizebox{2\columnwidth}{!}{
		\begin{tabular}{c|ccc|ccc||ccc|cc|cc|ccc||c}
			\hline
			\multicolumn{1}{c|}{Tracker} & \multicolumn{3}{c|}{GOT10k-test \cite{Huang2021}} & \multicolumn{3}{c||}{TrackingNet-test \cite{muller2018trackingnet}} & \multicolumn{3}{c|}{LaSOT-test \cite{fan2021lasot}} & \multicolumn{2}{c|} {NfS30 \cite{kiani2017need}} & \multicolumn{2}{c|} {UAV123 \cite{mueller2016benchmark}} & \multicolumn{3}{c||} {AVisT \cite{noman2022avist}} & \multicolumn{1}{c}{\textit{fps}}\\
			\multicolumn{1}{c|}{} & $AO$ & $SR_{0.50}$ & $SR_{0.75}$ & $AUC$ & $P_{norm}$ & $P$ & $AUC$ & $P_{norm}$ & $P$ & $AUC$ & $P$ & $AUC$ & $P$ & $AUC$ & OP$50$ & OP$75$ & (CPU)\\
			\hline
			LightTrack \cite{yan2021lighttrack} (CVPR'21) & 0.582 & 0.668 & 0.442 & 0.729 & 0.793 & 0.699 & 0.522 & 0.583 & 0.517 & 0.565 & 0.692 & 0.617 & 0.799 & 0.404 & 0.437 & 0.242 & 42 \\
			
			Stark-Lightning \cite{yan2021learning} (ICCV'21) & 0.596 & 0.696 & 0.479 & 0.727 & 0.779 & 0.674 & 0.578 & 0.660 & 0.574 & 0.596 & 0.710 & 0.620 & 0.820 & 0.394 & 0.431 & 0.223 & 50 \\
			
			FEAR-XS \cite{borsuk2022fear} (ECCV'22) & 0.573 & 0.681 & 0.455 & 0.715 & 0.805 & 0.699 & 0.501 & 0.594 & 0.523 & 0.486 & 0.563 & 0.610 & 0.816 & 0.370 & 0.421 & 0.220 & 42 \\
			
			HCAT \cite{chen2022efficient} (ECCV'22) & {\color{blue}0.634} & {\color{blue}0.743} & {\color{blue}0.558} & 0.763 & {\color{blue}0.824} & {\color{blue}0.726} & 0.590 & 0.683 & 0.605 & {\color{blue}0.619} & {\color{blue}0.741} & 0.620 & 0.805 & {\color{blue}0.418} & {\color{blue}0.481} & {\color{blue}0.263} & 45 \\
			
			E.T.Track \cite{blatter2023efficient} (WACV'23) & 0.566 & 0.646 & 0.425 & 0.740 & 0.798 & 0.698 & 0.589 & 0.670 & 0.603 & 0.570 & 0.694 & 0.626 & 0.808 & 0.390 & 0.412 & 0.227 & 44 \\
			
			MixFormerV2-S \cite{cui2023mixformerv2} (arXiv'23) & 0.587 & 0.672 & 0.482 & {\color{blue}0.767} & 0.812 & 0.714 & {\color{blue}0.610} & {\color{blue}0.694} & {\color{blue}0.614} & 0.610 & 0.722 & {\color{blue}0.634} & {\color{blue}0.837} & 0.396 & 0.425 & 0.227 & 30 \\
			
			SMAT (ours) & {\color{red}\textbf{0.645}} & {\color{red}\textbf{0.747}} & {\color{red}\textbf{0.578}} & {\color{red}\textbf{0.786}} & {\color{red}\textbf{0.842}} & {\color{red}\textbf{0.756}} & {\color{red}\textbf{0.617}} & {\color{red}\textbf{0.711}} & {\color{red}\textbf{0.646}} & {\color{red}\textbf{0.620}} & {\color{red}\textbf{0.746}} & {\color{red}\textbf{0.643}} & {\color{red}\textbf{0.839}} & {\color{red}\textbf{0.447}} & {\color{red}\textbf{0.507}} & {\color{red}\textbf{0.313}} & 37\\
			\hline
		\end{tabular}
	} 
	\caption{Comparison of proposed \textit{SMAT} with the related lightweight SN trackers on GOT10k-test (server), TrackingNet-test (server), LaSOT-test, NfS30, UAV123, and AVisT datasets. The best and second-best results are highlighted in {\color{red}red} and {\color{blue}blue}, respectively.}
	\label{table:qualitative_results}
\end{table*}
No related tracker performs consistently second best across the six datasets. HCAT exhibits the second-best results in 10 out of 16 metrics across all datasets, while MixFormerV2-S scores the second-best in 6 out of 16 cases. Regarding \textit{fps} under CPU, our tracker is relatively faster than MixFormerV2-S by 19\% and slower than HCAT by 17\%. Considering the GOT10k-test dataset (server-based evaluation), our tracker is better than MixFormerV2-S and HCAT on average by 3.5\% in $AO$, 4\% in $SR_{0.50}$, and 5.8\% in $SR_{0.75}$.

Since \textbf{GOT10k-test} dataset has videos with target object categories unseen during training, it is well-suited for evaluating tracker generalization. Our \textit{SMAT} results on the GOT10k-test demonstrate its superior generalization capability compared to the related lightweight trackers. Since E.T.Track has a transformer-based predictor head and MixFormerV2-S has a fully transformer-based backbone, these trackers are closely related to our work. By concurrently deploying a transformer-based backbone and head module, our \textit{SMAT} outperforms both E.T.Track and MixFormerV2 on average by a significant margin of 6.8\% in $AO$, 8.8\% in $SR_{0.50}$, and 12.4\% in $SR_{0.75}$. On \textbf{TrackingNet-test} benchmark, the proposed \textit{SMAT} has a better performance than all the related trackers by at least 1.9\% in $AUC$, 1.8\% in $P_{norm}$, and 3\% in $P$. No single tracker consistently exhibits the second-best performance on the TrackingNet dataset. The results of our \textit{SMAT} on the \textbf{LaSOT-test} videos show that our tracker has better long-term tracking performance than other lightweight trackers. The related E.T.Track employs a post-processing step to refine the predicted bounding boxes, which reduces the chances of target loss or drift during long-term tracking. The other related tracker, MixFormerV2, utilizes an online template update scheme to improve long-term tracking performance. Despite the absence of such bounding-box refinement step and template update scheme, our \textit{SMAT} performs better than E.T.Track by 2.8\% and MixFormerV2 by 0.7\% in $AUC$. The results on \textbf{NfS30} dataset show that our \textit{SMAT} is resilient to motion blur and is better-suited for tracking fast-moving objects than the related trackers. Similarly, proposed \textit{SMAT} has a better $AUC$ by at least 2.8\% and 0.9\% on \textbf{AVisT} and \textbf{UAV123} datasets, respectively, compared to the other lightweight trackers. It shows that our tracker performs better than the related trackers on videos affected by adverse visibility conditions and is robust to the challenges of airborne scenarios.

The last column of Table \ref{table:qualitative_results} lists the \textit{fps} of the proposed \textit{SMAT}
and related trackers, evaluated on a 12th Gen Intel(R) Core-i9 CPU. The proposed \textit{SMAT} tracker achieves a real-time speed of 37 \textit{fps} by leveraging the computational efficiency of the separable self and mixed attention blocks used in its model architecture. Compared to standard transformer-based MixFormerV2-S, our tracker is faster by 19\% since the proposed \textit{SMAT} uses separable attention-based transformers in its backbone for efficient computation of attention. However, in comparison to the related lightweight trackers with a two-stream pipeline \cite{yan2021lighttrack, yan2021learning, borsuk2022fear, chen2022efficient, blatter2023efficient}, our \textit{SMAT} has a 17\% lower \textit{fps} on average. It is mainly due to the coupling of features in our tracker backbone, requiring evaluation of template and search region features at every frame during inference. On the other hand, trackers with a two-stream pipeline compute the template features only once since they do not perform feature fusion in their backbone.

Our tracker achieves a speed of 158 \textit{fps} upon its evaluation on an Nvidia RTX 3090 GPU. With nearly 3.8 Million parameters, our model has a size of 15.3MB on disk.

\subsection{Ablation Study}
In this section, we present ablation study results quantifying the role of different components of our tracker. For this study, we use GOT10k-test \cite{Huang2021} and LaSOT-test \cite{fan2021lasot} due to their suitability towards gauging tracker generalization and long-term tracking performance, respectively. \\

\noindent {\bf Comparing feature fusion techniques:} In Section \ref{subsec:backbone}, we mentioned that the proposed mixed attention efficiently approximates the explicit modeling of interaction \textit{within} and \textit{between} the target template and search regions. In this section, we experimentally verify the efficacy and efficiency of the proposed approach by comparing its results with other feature fusion methods that deploy explicit computation of self or cross-attention (or both).  
\begin{figure}[t]
	\centering
	\includegraphics[width=\linewidth]{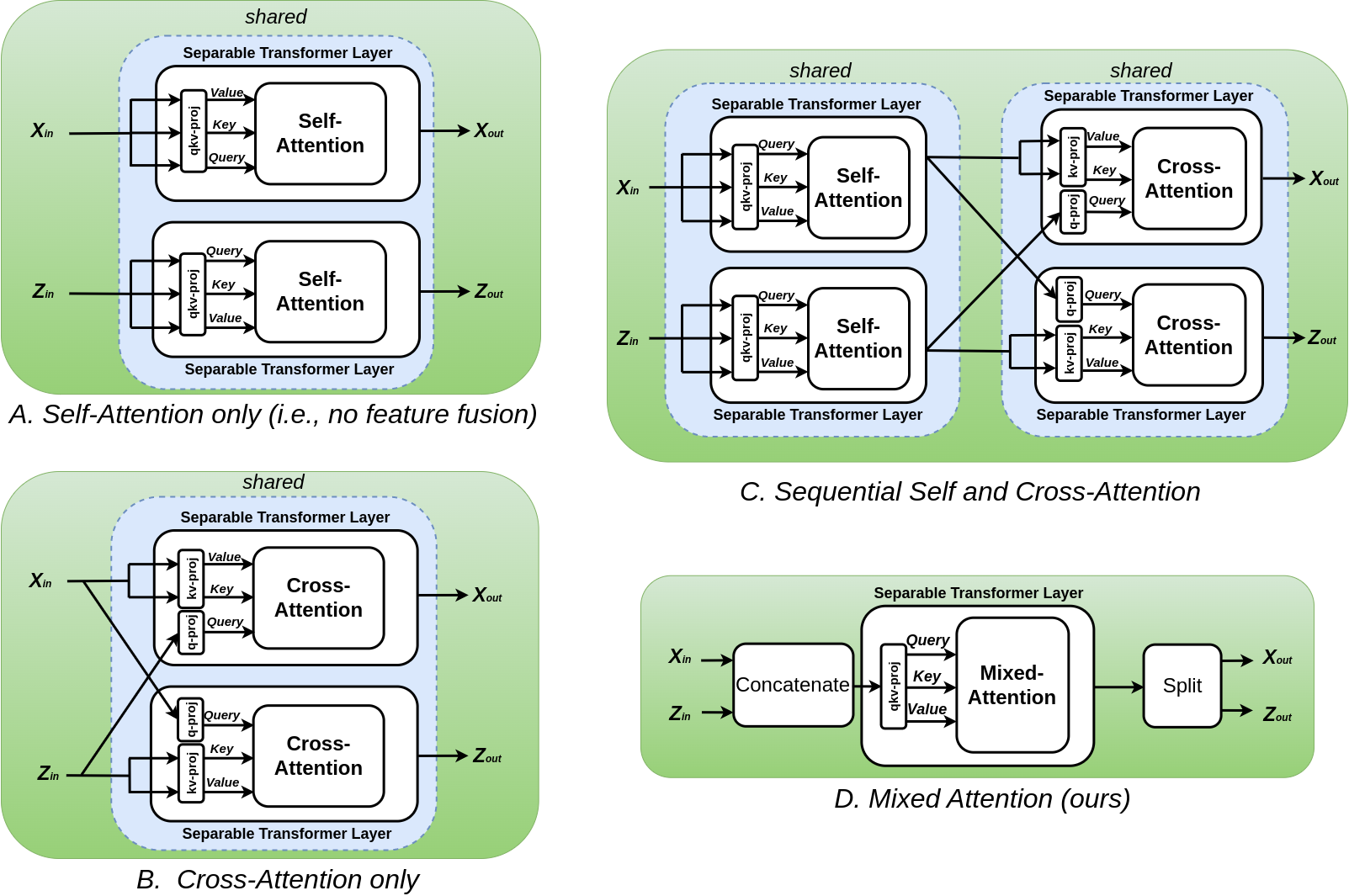}
	\caption{Comparing different feature fusion techniques (\textit{A, B} and \textit{C}) with the proposed mixed attention-based method shown in \textit{D}.} 
	\label{fig:attn_ablation}
\end{figure}
For the variant-\textit{A} shown in Figure \ref{fig:attn_ablation}, we use a shared separable self-attention transformer block for the template and search regions. This approach facilitates independent computation of the template and search features for high tracking speed; however, it restricts the information exchange between the two regions (i.e., no feature fusion). For the variant-\textit{B}, we deploy shared separable cross-attention transformers for inter-region feature fusion but no intra-region feature modeling (i.e., no self-attention). A similar cross-feature blending approach for relation modeling has shown excellent tracking results \cite{chen2021transformer}. Lastly, we implement cascaded self and cross-attention transformers for the third variant-\textit{C} to explicitly model inter and intra-region feature fusion. The proposed mixed attention block, shown as variant-\textit{D} in Figure \ref{fig:attn_ablation}, approximates explicit self and cross-attention computation by applying a separable transformer block on the concatenated features from the template and search regions.
\begin{table}[t]
	\centering
	\resizebox{\columnwidth}{!}{
		\begin{tabular}{c|ccc|ccc|c}
			\hline
			\multicolumn{1}{c|}{Attention} & \multicolumn{3}{c|}{GOT10k-test \cite{Huang2021}} & \multicolumn{3}{c|}{LaSOT-test \cite{fan2021lasot}} & \multicolumn{1}{c}{\textit{fps}} \\
			\multicolumn{1}{c|}{Mechanism} & $AO$ & $SR_{0.50}$ & $SR_{0.75}$ & $AUC$ & $P_{norm}$ & $P$ & (CPU)\\
			\hline
			
			\textit{A} & 0.631 & 0.726 & 0.578 & 0.604 & 0.689 & 0.629 & {\color{red}\textbf{39}} \\
			
			\textit{B} & {\color{blue}0.645} & 0.743 & {\color{blue}0.590} & 0.609 & 0.696 & 0.631 & 30 \\
			
			\textit{C} & {\color{red}\textbf{0.654}} & {\color{red}\textbf{0.761}} & {\color{red}\textbf{0.598}} & {\color{red}\textbf{0.621}} & {\color{red}\textbf{0.717}} & {\color{red}\textbf{0.656}} & 24 \\
			
			\textit{D} (ours) & {\color{blue}0.645} & {\color{blue}0.747} & 0.578 & {\color{blue}0.617} & {\color{blue}0.711} & {\color{blue}0.646}  & {\color{blue}37}\\
			\hline
		\end{tabular}
	} 
	\caption{Summarizing the feature fusion-based ablation study results for the proposed \textit{SMAT} tracker. The best and second-best results are highlighted in {\color{red}red} and {\color{blue}blue}, respectively.}
	\label{table:feature_fusion_ablation_study_results}
\end{table}
Table \ref{table:feature_fusion_ablation_study_results} summarizes the performance of these variants on GOT10k and LaSOT test datasets. From Table \ref{table:feature_fusion_ablation_study_results}, we can see that variant-\textit{A} has a $1.05 \times$ higher \textit{fps} than the proposed method; however, the lack of information flow between the template and search region impacts its performance. Hence, compared to the proposed tracker, it has a lower $AUC$ score by 1.4\% and 1.3\% on GOT10k and LaSOT, respectively. Performance of the pure cross-attention-based variant-\textit{B} is lower than our \textit{SMAT} by 0.8\% in $AUC$ on the LaSOT dataset, and both approaches have comparable performance on the GOT10k dataset. However, variant-\textit{B} has a relatively lower \textit{fps} by 18.9\% than our \textit{SMAT}, which indicates that separately computing cross-attention for the template and search regions is slower than our mixed-attention computation on concatenated features. Variant-\textit{C} achieves the best results on both datasets; however, explicit computation of self and cross-attention significantly impacts its tracking speed. Therefore, it has the lowest \textit{fps} value compared to the other variants and is relatively slower than our approach by 35\% on a CPU. Compared to the variant-\textit{C}, our method approximates the inter and intra-region feature fusion by a single attention operation. Therefore, as seen from Table \ref{table:feature_fusion_ablation_study_results}, our mixed attention-based approach provides the best trade-off between performance and speed compared to other feature fusion variants shown in Figure \ref{fig:attn_ablation}. \\

\noindent {\bf Convolutional vs Transformer-based head:} To quantify the significance of our separable self-attention transformer-based predictor head from Section \ref{subsec:predictor_head}, we replace the proposed module with a fully-convolutional predictor head for classification and bounding-box regression. As seen from Table \ref{table:predictor_head_ablation_study_results}, this replacement decreases the performance of the proposed \textit{SMAT} tracker by 3.5\% in $AO$ for the GOT10k-test dataset and 2.3\% in $AUC$ for the LaSOT-test dataset, highlighting the impact of global contextual modeling of encoded features by the proposed predictor head. \\
%
%
\begin{table}[t]
	\centering
	\resizebox{\columnwidth}{!}{
		\begin{tabular}{c|ccc|ccc}
			\hline
			\multicolumn{1}{c|}{Predictor Head} & \multicolumn{3}{c|}{GOT10k-test \cite{Huang2021}} & \multicolumn{3}{c}{LaSOT-test \cite{fan2021lasot}}\\
			\multicolumn{1}{c|}{} & $AO$ & $SR_{0.50}$ & $SR_{0.75}$ & $AUC$ & $P_{norm}$ & $P$ \\
			\hline
			
			Fully-Convolutional & 0.610 & 0.709 & 0.540 & 0.594 & 0.682 & 0.612 \\
			
			Transformer-based (ours) & {\color{red}\textbf{0.645}} & {\color{red}\textbf{0.747}} & {\color{red}\textbf{0.578}} & {\color{red}\textbf{0.617}} & {\color{red}\textbf{0.711}} & {\color{red}\textbf{0.646}}  \\
			\hline
		\end{tabular}
	} 
	\caption{Ablation study results for the proposed transformer-based prediction head. Best results are highlighted in {\color{red}red}.}
	\label{table:predictor_head_ablation_study_results}
\end{table}

\noindent {\bf Standard vs Separable Attention mechanism:} To evaluate the efficiency of the separable attention mechanism deployed in the proposed \textit{SMAT}, we retrain our model by replacing the separable transformer blocks in the tracker backbone with the standard transformer-based MobileViT block \cite{mehta2022mobilevit}.
\begin{table}[t]
	\centering
	\resizebox{\columnwidth}{!}{
		\begin{tabular}{c|ccc|ccc||c}
			\hline
			\multicolumn{1}{c|}{Mixed Attention} & \multicolumn{3}{c|}{GOT10k-test \cite{Huang2021}} & \multicolumn{3}{c||}{LaSOT-test \cite{fan2021lasot}} & \multicolumn{1}{c}{\textit{fps}}\\
			\multicolumn{1}{c|}{Mechanism} & $AO$ & $SR_{0.50}$ & $SR_{0.75}$ & $AUC$ & $P_{norm}$ & $P$ & (CPU)\\
			\hline
			
			Standard & {\color{red}\textbf{0.659}} & {\color{red}\textbf{0.760}} & {\color{red}\textbf{0.605}} & 0.600 & 0.687 & 0.630 & 32\\
			
			Separable (ours) & 0.645 & 0.747 & 0.578 & {\color{red}\textbf{0.617}} & {\color{red}\textbf{0.711}} & {\color{red}\textbf{0.646}} & {\color{red}\textbf{37}} \\
			\hline
		\end{tabular}
	} 
	\caption{Comparison of tracker performance using the standard \textit{vs} separable mixed attention mechanism in the backbone. The best results are highlighted in {\color{red}red}.}
	\label{table:standard_vs_separable_attention_ablation_study_results}
\end{table}
From Table \ref{table:standard_vs_separable_attention_ablation_study_results}, we observe that the proposed method has a higher $AUC$ of 1.7\% on the LaSOT dataset compared to the standard attention-based tracker. On the other hand, our approach has a lower $AUC$ of 1.4\% on the GOT10k dataset. However, the efficient attention evaluation enhances the speed of the proposed tracking approach by 15.6\% compared to the standard transformer-based tracking, as seen from Table \ref{table:standard_vs_separable_attention_ablation_study_results}.

\subsection{Attribute-based analysis} \label{sec:attrib_analysis}
\begin{table*}[t]
	\centering
	\resizebox{2\columnwidth}{!}{
		\begin{tabular}{c|c|c|c|c|c|c|c|c|c|c|c|c|c|c||c}
			\hline
			\multicolumn{1}{c|}{Tracker} & \multicolumn{1}{c|}{\textit{ARC}} & \multicolumn{1}{c|}{\textit{BC}} & \multicolumn{1}{c|}{\textit{CM}} & \multicolumn{1}{c|}{\textit{DEF}} & \multicolumn{1}{c|}{\textit{FM}} & \multicolumn{1}{c|}{\textit{FOC}} & \multicolumn{1}{c|}{\textit{IV}} & \multicolumn{1}{c|}{\textit{LR}} & \multicolumn{1}{c|}{\textit{MB}} & \multicolumn{1}{c|}{\textit{OV}} & \multicolumn{1}{c|}{\textit{POC}} & \multicolumn{1}{c|}{ROT} & \multicolumn{1}{c|}{SV} & \multicolumn{1}{c||}{\textit{VC}} & \multicolumn{1}{c}{Overall}\\
			\hline
			
			LightTrack \cite{yan2021lighttrack} & 0.503 & 0.434 & 0.539 & 0.577 & 0.334 & 0.386 & 0.550 & 0.407 & 0.457 & 0.441 & 0.497 & 0.519 & 0.523 & 0.502 & 0.522\\
			
			Stark-Lightning \cite{yan2021learning} & 0.572 & 0.491 & 0.613 & 0.594 & {\color{blue}0.471} & 0.505 & 0.610 & 0.516 & 0.568 & 0.557 & 0.554 & 0.577 & 0.582 & 0.581 & 0.578 \\
			
			FEAR-XS \cite{borsuk2022fear} & 0.488 & 0.437 & 0.528 & 0.505 & 0.389 & 0.403 & 0.506 & 0.421 & 0.473 & 0.425 & 0.478 & 0.489 & 0.506 & 0.487 & 0.501\\
			
			HCAT \cite{chen2022efficient} & 0.587 & 0.524 & 0.639 & 0.619 & 0.460 & 0.507 & 0.606 & 0.520 & 0.579 & 0.538 & 0.568 & 0.592 & 0.600 & 0.567 & 0.590 \\
			
			E.T.Track \cite{blatter2023efficient} & 0.573 & {\color{blue}0.526} & 0.590 & 0.619 & 0.404 & 0.480 & 0.612 & 0.484 & 0.545 & 0.519 & 0.562 & 0.588 & 0.594 & 0.576 & 0.589\\
			
			MixFormerV2-S \cite{cui2023mixformerv2} & {\color{blue}0.603} & 0.519 & {\color{blue}0.642} & {\color{blue}0.626} & {\color{red}0.507} & {\color{red}0.539} & {\color{blue}0.619} & {\color{red}0.556} & {\color{red}0.604} & {\color{red}0.574} & {\color{blue}0.586} & {\color{blue}0.603} & {\color{blue}0.617} & {\color{red}0.630} & {\color{blue}0.610}\\
			
			SMAT (ours) & {\color{red}0.610} & {\color{red}0.553} & {\color{red}0.656} & {\color{red}0.663} & 0.466 & {\color{blue}0.517} & {\color{red}0.662} & {\color{blue}0.523} & {\color{blue}0.592} & {\color{blue}0.570} & {\color{red}0.600} & {\color{red}0.621} & {\color{red}0.624} & {\color{blue}0.597} & {\color{red}0.617}\\
			
			\hline
		\end{tabular}
	} 
	\caption{Comparing the $AUC$ values of the proposed \textit{SMAT} with the related lightweight trackers for 14 attributes of the LaSOT dataset. The best and second-best results are highlighted in {\color{red}red} and {\color{blue}blue}, respectively. The last column indicates the mean $AUC$ across all videos.}
	\label{table:per_attribute_results}
\end{table*}
In this section, we compare the per-attribute performance of the proposed \textit{SMAT} tracker with related lightweight trackers on the LaSOT dataset. Table \ref{table:per_attribute_results} summarizes the tracker evaluation results on various challenging factors (or attributes) of the LaSOT dataset, namely Aspect Ratio Change (\textit{ARC}), Background Clutter (\textit{BC}), Camera Motion (\textit{CM}), Deformation (\textit{DEF}), Fast Motion (\textit{FM}), Full Occlusion (\textit{FOC}), Illumination Variation (\textit{IV}), Low Resolution (\textit{LR}), Motion Blur (\textit{MB}), Out-of-View (\textit{OV}), Partial Occlusion (\textit{POC}), Rotation (\textit{ROT}), Scale Variation (\textit{SV}), and Viewpoint Change (\textit{VC}).

From Table \ref{table:per_attribute_results}, we can see that our tracker has the best performance on 8 out of 14 attributes, and it has the second-best performance in 5 cases. For the attributes \textit{IV} and \textit{DEF}, the proposed \textit{SMAT} performs significantly better than the second-best tracker MixFormerV2-S\cite{cui2023mixformerv2}, with a higher $AUC$ of 4.3\% and 3.7\%, respectively. In comparison to the related trackers that do not perform feature fusion in their backbone, i.e., \cite{yan2021lighttrack, yan2021learning, borsuk2022fear, chen2022efficient, blatter2023efficient}, our tracker has a higher $AUC$ of 4.4\% for \textit{DEF} and 2.3\% for \textit{ARC} than the second best trackers, \cite{chen2022efficient, blatter2023efficient} and \cite{chen2022efficient}, respectively. It indicates the effectiveness of transformer-based feature fusion in our tracker backbone, producing accurate bounding boxes under drastic target appearance variations. Also, in comparison to these trackers, our \textit{SMAT} is resilient to tracking failures under \textit{POC} and \textit{BC}, with a higher $AUC$ of 3.2\% and 2.7\%, respectively, than the second-best results. On the other hand, our \textit{SMAT} has an inferior performance than the related trackers \cite{cui2023mixformerv2} and \cite{yan2021learning} for the attribute \textit{FM}; we are working to improve our \textit{SMAT} performance under \textit{FM}.
%

\subsection{Visualizing the attention maps}
\begin{figure}[t]
	\centering
	\includegraphics[width=0.95\linewidth]{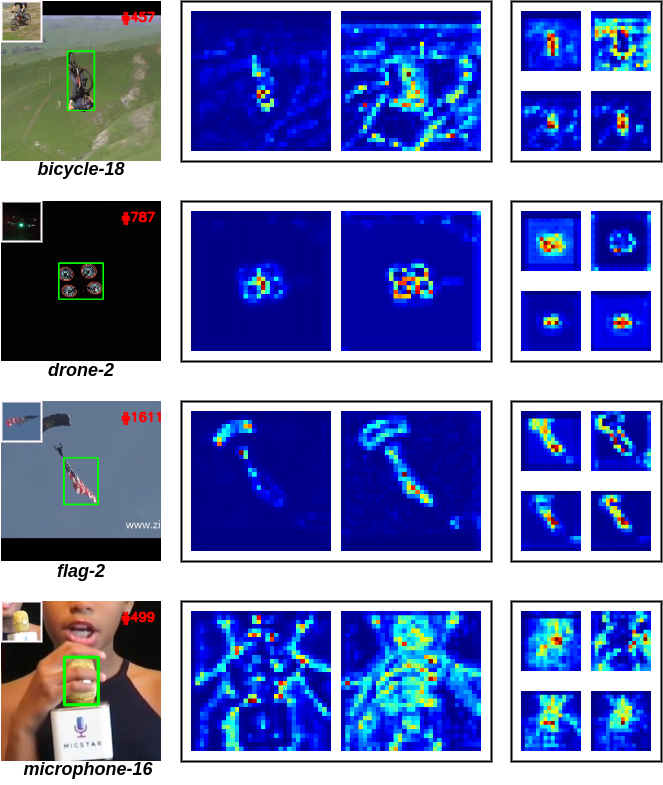}
	\caption{Visualizing the bounding box output (left) and the corresponding attention maps (center and right) for the proposed \textit{SMAT} tracker. The larger values in the attention map are denoted by red color, while the smaller values are represented by blue color.}
	\label{fig:attn_visualization}
\end{figure}

To showcase the interpretability of the proposed \textit{SMAT} tracker, we visualize the tracker output and the corresponding attention maps in Figure \ref{fig:attn_visualization} for four sequences chosen from the LaSOT test dataset. The images on the left contain the target template (top-left corner), the search region at frame $\#t$, and the tracker output. The images in the center and right indicate the attention maps corresponding to the transformer blocks of our tracker backbone at the spatial resolution of $32 \times 32$ and $16 \times 16$, respectively. For the examples shown in Figure \ref{fig:attn_visualization}, the target object is impacted by a challenging factor described in Section \ref{sec:attrib_analysis}, i.e., \textit{ARC} for \textit{bicycle-18}, \textit{IV} for \textit{drone-2}, \textit{DEF} for \textit{drone-2}, and \textit{POC} for \textit{microphone-16}. Despite the influence of these attributes, our \textit{SMAT} successfully locates the target object. For the example shown in the last row of Figure \ref{fig:attn_visualization}, the target is partially occluded by an external object. In this case, the tracker focuses on the visual cues around the target object, as seen from the attention maps in the center. This information is processed by the subsequent transformer blocks of our tracker backbone to produce stronger attention values in the target center and generate an accurate bounding box.

\section{Conclusion}
This paper proposed a separable self and mixed attention transformer-based architecture for lightweight tracking. The proposed backbone utilized the separable mixed attention transformer layer to facilitate the exchange of information between the target template and the search region and generate improved encoding compared to the two-stream tracking pipeline. The proposed separable self-attention transformer-based predictor head efficiently modeled long-range dependencies within the fused encoding to generate superior target classification and bounding-box prediction results. Our ablation study analyzed the accuracy-speed tradeoffs using different feature fusion methods, showcased the effectiveness of the proposed head module for accurate tracking, and demonstrated the efficiency of the separable mixed attention compared to the standard attention-based tracking. Our \textit{SMAT} performed better than related lightweight trackers on six challenging benchmarks. The computational efficiency of the proposed architecture assisted our tracker, with 3.8M parameters, to exceed real-time speed on a CPU, while running at 158 \textit{fps} on GPU.

{\small
\bibliographystyle{ieee_fullname}
\bibliography{egbib,strings,Goutamrefs}
}

\end{document}